# ECO: Efficient Convolutional Network for Online Video Understanding

Mohammadreza Zolfaghari, Kamaljeet Singh and Thomas Brox

University of Freiburg
{zolfagha,singhk,brox}@cs.uni-freiburg.de

**Abstract.** The state of the art in video understanding suffers from two problems: (1) The major part of reasoning is performed locally in the video, therefore, it misses important relationships within actions that span several seconds. (2) While there are local methods with fast per-frame processing, the processing of the whole video is not efficient and hampers fast video retrieval or online classification of long-term activities. In this paper, we introduce a network architecture[1] that takes long-term content into account and enables fast per-video processing at the same time. The architecture is based on merging long-term content already in the network rather than in a post-hoc fusion. Together with a sampling strategy, which exploits that neighboring frames are largely redundant, this yields high-quality action classification and video captioning at up to 230 videos per second, where each video can consist of a few hundred frames. The approach achieves competitive performance across all datasets while being 10x to 80x faster than state-of-the-art methods.

## 1 Introduction

Video understanding and, specifically, action classification have benefited a lot from deep learning and the larger datasets that have been created in recent years. The new datasets, such as Kinetics [1], ActivityNet [2], and Something-Something [3] have contributed more diversity and realism to the field. Deep learning provides powerful classifiers at interactive frame rates, enabling applications like real-time action detection [4].

While action detection, which quickly decides on the present action within a short time window, is fast enough to run in real-time, activity understanding, which is concerned with longer-term activities that can span several seconds, requires the integration of the long-term context to achieve full accuracy. Several 3D CNN architectures have been proposed to capture temporal relations between frames, but they are computationally expensive and, thus, can cover only comparatively small windows rather than the entire video. Existing methods typically use some post-hoc integration of window-based scores, which is suboptimal for exploiting the temporal relationships between the windows.

[1] https://github.com/mzolfaghari/ECO-efficient-video-understanding

In this paper, we introduce a straightforward, end-to-end trainable architecture that exploits two important principles to avoid the above-mentioned dilemma. Firstly, a good initial classification of an action can already be obtained from just a single frame. The temporal neighborhood of this frame comprises mostly redundant information and is almost useless for improving the belief about the present action[2]. Therefore, we process only a single frame of a temporal neighborhood efficiently with a 2D convolutional architecture in order to capture appearance features of such frame. Secondly, to capture the contextual relationships between distant frames, a simple aggregation of scores is insufficient. Therefore, we feed the feature representations of distant frames into a 3D network that learns the temporal context between these frames and so can improve significantly over the belief obtained from a single frame – especially for complex long-term activities. This principle is much related to the so-called early or late fusion used for combining the RGB stream and the optical flow stream in two-stream architectures. However, this principle has been mostly ignored so far for aggregation over time and is not part of the state-of-the-art approaches.

Consequent implementation of these two principles together leads to a high classification accuracy without bells and whistles. The long temporal context of complex actions can be fully captured, whereas the fact that the method only looks at a very small fraction of all frames in the video leads to extremely fast processing of entire videos. This is very beneficial especially in video retrieval applications.

Additionally, this approach opens the possibility for online video understanding. In this paper, we also present a way to use our architecture in an online setting, where we provide a fast first guess on the action and refine it using the longer term context as a more complex activity establishes. In contrast to online action detection, which has been enabled recently [4], the approach provides not only fast reaction times, but also takes the longer term context into account.

We conducted experiments on various video understanding problems including action recognition and video captioning. Although we just use RGB images as input, we obtain on-par or favorable performance compared to state-of-the-art approaches on most datasets. The runtime-accuracy trade-off is superior on all datasets.

## 2 Related Work

**Video classification with deep learning.** Most recent works on video classification are based on deep learning [6,7,8,9,10]. To explore the temporal context of a video, 3D convolutional networks are on obvious option. Tran et al. [11] introduced a 3D architecture with 3D kernels to learn spatio-temporal features from a sequence of frames. In a later work, they studied the use of a Resnet architecture with 3D convolutions and showed the improvements over their earlier c3d architecture [9]. An alternative way to model the temporal relation between frames is by using recurrent networks [6,12,13]. Donahue et al. [6] employed a LSTM to integrate features from a CNN over time. However, the performance of recurrent networks on action recognition currently lags behind that of recent CNN-based methods, which may indicate that they do not sufficiently model long-term dynamics [12,13]. Recently, several works utilized 3D architectures for action recognition [14,10,15]. These approaches model the short-term temporal context of the input video based on a sliding window. At inference time, these methods must compute the average score over multiple windows, which is quite time consuming. For example, ARTNet [15] requires on average 250 samples to classify one video.

[2] An exception is the use of two frames for capturing motion, which could be achieved by optionally feeding optical flow together with the RGB image. In this paper, we only provide RGB images, but an extension with optical flow, e.g., a fast variant of FlowNet [5] would be straightforward.

All these approaches do not sufficiently use the comprehensive information from the entire video during training and inference. Partial observation not only causes confusion in action prediction, but also requires an extra post-processing step to fuse scores. Extra feature/score aggregation reduces the speed of video processing and disables the method to work in a real-time setting.

**Long-term representation learning.** To cope with the problem of partial observation, some methods increased the temporal resolution of the sliding window [16,17]. However, expanding the temporal length of the input has two major drawbacks. (1) It is computationally expensive, and (2) still fails to cover the visual information of the entire video, especially for longer videos.

Some works proposed encoding methods [18,19,20] to learn a video representation from samples. In these approaches, features are usually calculated for each frame independently and are aggregated across time to make a video-level representation. This ignores the relationship between the frames.

TSN [21] employed a sparse and global temporal sampling method to choose frames from the entire video during training. However, as in the aggregation methods above, frames are processed independently at inference time and their scores are aggregated only in the end. Consequently, the performance in their experiments stays the same when they change the number of samples, which indicates that their model does not really benefit from the long-range temporal information.

Our work is different from these previous approaches in three main aspects: (1) Similar to TSN, we sample a fixed number of frames from the entire video to cover long-range temporal structure for understanding of video. In this way, the sampled frames span the entire video independent of the length of the video. (2) In contrast to TSN, we use a 3D-network to learn the relationship between the frames and track them throughout the video. The network is trained end-to-end to learn this relationship. (3) The network directly provides video-level scores without post-hoc feature aggregation. Therefore, it can be run in online mode and in real-time even on small computing devices.

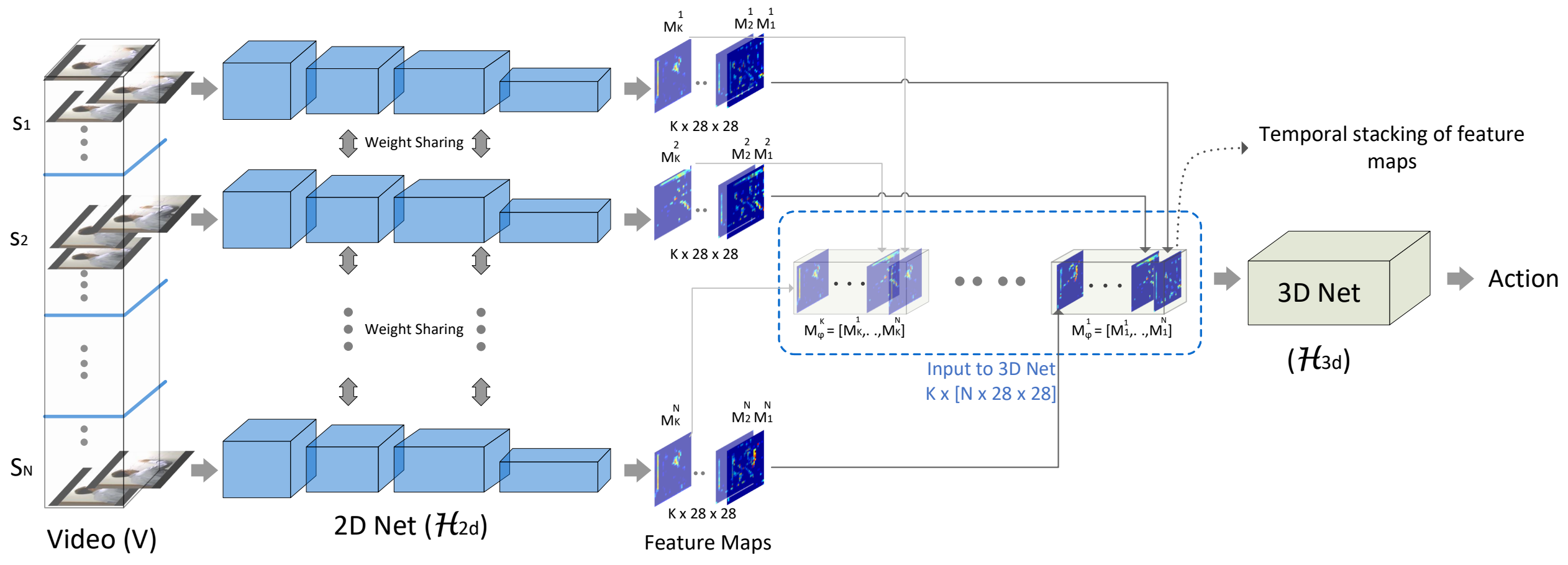

Fig. 1: **Architecture overview of ECO Lite.** Each video is split into $N$ subsections of equal size. From each subsection a single frames is randomly sampled. The samples are processed by a regular 2D convolutional network to yield a representation for each sampled frame. These representations are stacked and fed into a 3D convolutional network, which classifies the action, taking into account the temporal relationship.

**Video Captioning.** Video captioning is a widely studied problem in computer vision [22,23,24,25]. Most approaches use a CNN pre-trained on image classification or action recognition to generate features [25,24,23]. These methods, like the video understanding methods described above, utilize a frame-based feature aggregation (e.g. Resnet or TSN) or a sliding window over the whole video (e.g. C3D) to generate video-level features. The features are then passed to a recurrent neural network (e.g. LSTM) to generate the video captions via a learned language model. The extracted visual features should represent both the temporal structure of the video and the static semantics of the scene. However, most approaches suffer from the problem that the temporal context is not properly extracted. With the network model in this work, we address this problem, and can consequently improve video captioning results.

**Real-time and online video understanding.** Deep learning accelerated image classification, but video classification remains challenging in terms of speed. A few works dealt with real-time video understanding [26,27,4,28]. EMV [27] introduced an approach for fast calculation of motion vectors. Despite this improvement, video processing is still slow. Kantorov [26] introduced a fast dense trajectory method. The other works used frame-based hand-crafted features for online action recognition [29,30]. Both accuracy and speed of feature extraction in these methods are far from that of deep learning methods. Soomro et al. [28] proposed an online action localization approach. Their model utilizes an expensive segmentation method which, therefore, cannot work in real-time. More recently, Singh et al. [4] proposed an online detection approach based on frame-level detections at 40fps. We compare to the last two approaches in Section 5.

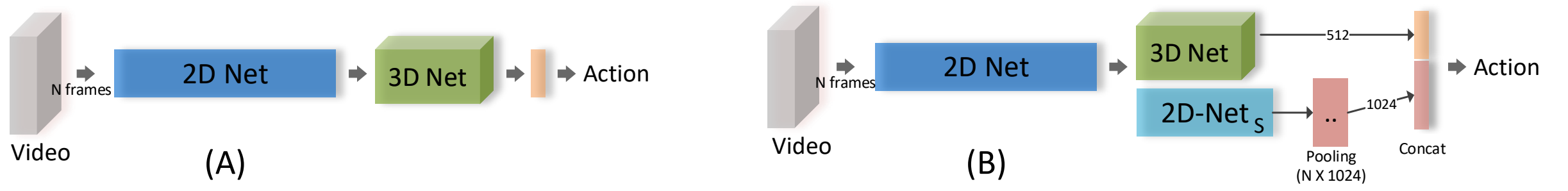

Fig. 2: **(A)** ECO Lite architecture as shown in more detail in Fig. 1. **(B)** Full ECO architecture with a parallel 2D and 3D stream.

## 3 Long-term Spatio-temporal Architecture

The network architecture is shown in Fig. 1. A whole video with a variable number of frames is provided as input to the network. The video is split into $N$ subsections $S_i$, $i = 1, ..., N$ of equal size, and in each subsection, exactly one frame is sampled randomly. Each of these frames is processed by a single 2D convolutional network (weight sharing), which yields a feature representation encoding the frames appearance. By jointly processing frames from time segments that cover the whole video, we make sure that we capture the most relevant parts of an action over time and the relationship among these parts.

Randomly sampling the position of the frame is advantageous over always using the same position, because it leads to more diversity during training and makes the network adapt to variations in the instantiation of an action. Note that this kind of processing exploits all frames of the video during training to model the variation. At the same time, the network must only process $N$ frames at runtime, which makes the approach very fast. We also considered more clever partitioning strategies that take the content of the subsections into account. However, this comes with the drawback that each frame of the video must be processed at runtime to obtain the partitioning, and the actual improvement by such smarter partitioning is limited, since most of the variation is already captured by the random sampling during training.

Up to this point, the different frames in the video are processed independently. In order to learn how actions are made up of the different appearances of the scene over time, we stack the representations of all frames and feed them into a 3D convolutional network. This network yields the final action class label.

The architecture is very straightforward, and it is obvious that it can be trained efficiently end-to-end directly on the action class label and on large datasets. It is also an architecture that can be easily adapted to other video understanding tasks, as we show later in the video captioning section 5.4.

### 3.1 ECO Lite and ECO Full

The 3D architecture in ECO Lite is optimized for learning relationships between the frames, but it tends to waste capacity in case of simple short-term actions that can be recognized just from the static image content. Therefore, we suggest an extension of the architecture by using a 2D network in parallel; see Fig.2(B). For the simple actions, this 2D network architecture can simplify processing and ensure that the static image features receive the necessary importance, whereas the 3D network architecture takes care of the more complex actions that depend on the relationship between frames.

The 2D network receives feature maps of all samples and produces $N$ feature representations. Afterwards, we apply average pooling to get a feature vector that is a representative for static scene semantics. We call the full architecture ECO and the simpler architecture in Fig. 2(A) ECO Lite.

### 3.2 Network details

**2D-Net:** For the 2D network ($\mathcal{H}_{2D}$) that analyzes the single frames, we use the first part of the BN-Inception architecture (until inception-3c layer) [31]. Details are given in the supplemental material. It has 2D filters and pooling kernels with batch normalization. We chose this architecture due to its efficiency. The output of $\mathcal{H}_{2D}$ for each single frame consist of 96 feature maps with size of $28 \times 28$.

**3D-Net:** For the 3D network $\mathcal{H}_{3D}$ we adopt several layers of 3D-Resnet18 [32], which is an efficient architecture used in many video classification works [32,15]. Details on the architecture are provided in the supplemental material. The output of $\mathcal{H}_{3D}$ is a one-hot vector for the different class labels.

**2D-Net$_S$ :** In the ECO full design, we use 2D-$Net_s$ in parallel with 3D-net to directly providing static visual semantics of video. For this network, we use the BN-Inception architecture from inception-4a layer until last pooling layer [31]. The last pooling layer will produce 1024 dimensional feature vector for each frame. We apply average pooling to generate video-level feature and then concatenate with features obtained from 3D-net.

### 3.3 Training details

We train our networks using mini-batch SGD with Nesterov momentum and utilize dropout in each fully connected layer. We split each video into $N$ segments and randomly select one frame from each segment. This sampling provides robustness to variations and enables the network to fully exploit all frames. In addition, we apply the data augmentation techniques introduced in [33]: we resize the input frames to $240 \times 320$ and employ fixed-corner cropping and scale jittering with horizontal flipping (temporal jittering provided by sampling). Afterwards, we run per-pixel mean subtraction and resize the cropped regions to $224 \times 224$.

The initial learning rate is 0.001 and decreases by a factor of 10 when validation error saturates for 4 epochs. We train the network with a momentum of 0.9, a weight decay of 0.0005, and mini-batches of size 32.

We initialize the weights of the 2D-Net weights with the BN-Inception architecture [31] pre-trained on Kinetics, as provided by [33]. In the same way, we use the pre-trained model of 3D-Resnet-18, as provided by [15] for initializing the weights of our 3D-Net. Afterwards, we train ECO and ECO Lite on the Kinetics dataset for 10 epochs.

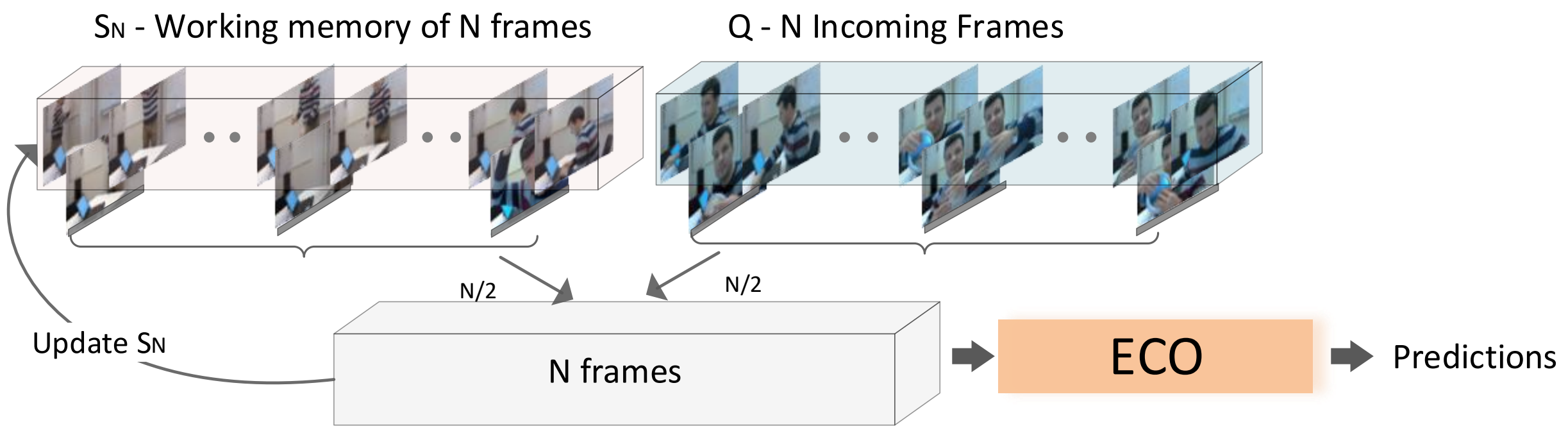

Fig. 3: Scheme of our sampling strategy for online video understanding. Half of the frames are sampled uniformly from the working memory in the previous time step, the other half from the queue (Q) of incoming frames.

For other datasets, we finetune the above ECO/ECO Lite models on the new datasets. Due to the complexity of the Something-Something dataset, we finetune the network for 25 epochs reducing the learning rate every 10 epochs by a factor of 10. For the rest, we finetune for 4k iterations and the learning rate drops by a factor of 10 as soons as the validation loss saturates. The whole training process on UCF101 and HMDB51 takes around 3 hours on one Tesla P100 GPU for the ECO architecture. We adjusted the dropout rate and the number of iterations based on the dataset size.

### 3.4 Test time inference

Most state-of-the-art methods run some post-processing on the network result. For instance, TSN and ARTNet [33,15], collect 25 independent frames/volumes per video, and for each frame/volume sample 10 regions by corner and center cropping, and their horizontal flipping. The final prediction is obtained by averaging the scores of all 250 samples. This kind of inference at test time is computationally expensive and thus unsuitable for real-time setting.

In contrast, our network produces action labels for the whole video directly without any additional aggregation. We sample $N$ frames from the video, apply only center cropping then feed them directly to the network, which provides the prediction for the whole video with a single pass.

## 4 Online video understanding

Most works on video understanding process in batch mode, i.e., they assume that the whole video is available when processing starts. However, in several application scenarios, the video will be provided as a stream and the current belief is supposed to be available at any time. Such online processing is possible with a sliding window approach, yet this comes with restrictions regarding the size of the window, i.e., long-term context is missing, or with a very long delay.

**Algorithm 1:** Online video understanding

**Input** : Live video stream ($V$), ECO pretrained model ($ECO_{NF}$), Number of Samples =Sampling window ($N$)
**Output:** Predictions
Initialize an empty queue $Q$ to queue N incoming frames;
Initialize working memory $S_N$ ;
Initialize average predictions $P_A$;
**while** *new frames available from* $V$ **do**
  Add frame $f_i$ from $V$ to queue $Q$;
  **if** $i$ % $N$ **then**
    $S_N$ := Sample 50% frames $Q$ and 50% from $S_N$ ;
    Empty queue $Q$;
    Feed $S_N$ to model $ECO_{NF}$ to get output probabilities $P$;
    $P_A$ := Average $P$ and $P_A$ ;
    Output average predictions $P_A$;
  **end**
**end**

In this section, we show how ECO can be adapted to run very efficiently in online mode, too. The modification only affects the sampling part and keeps the network architecture unchanged. To this end, we partition the incoming video content into segments of $N$ frames, where $N$ is also the number of frames that go into the network. We use a working memory $S_N$, which always comprises the $N$ samples that are fed to the network together with a time stamp. When a video starts, i.e., only $N$ frames are available, all $N$ frames are sampled densely and are stored in the working memory $S_N$. With each new time segment, $N$ additional frames come in, and we replace half of the samples in $S_N$ by samples from this time segment and update the prediction of the network; see Fig. 3. When we replace samples from $S_N$, we uniformly replace samples from previous time segments. This ensures that changes can be anticipated in real time, while the temporal context is taken into account and slowly fades out via the working memory. Details on the update of $S_N$ are shown in Algorithm 1.

The online approach with ECO runs at 675 fps (and at 970 fps with ECO Lite) on a Tesla P100 GPU. In addition, the model is memory efficient by just keeping exactly N frames. This enables the implementation also on much smaller hardware, such as mobile devices. The video in the supplemental material shows the recorded performance of the online version of ECO in real-time.

## 5 Experiments

We evaluate our approach on different video understanding problems to show the generalization ability of approach. We evaluated the network architecture on the most common action classification datasets in order to compare its performance against the state-of-the-art approaches. This includes the older but still very popular datasets UCF101 [34] and HMDB51 [35], but also the more recent datasets Kinetics [1] and Something-Something [3]. Moreover, we applied the architecture to video captioning and tested it on the widely used Youtube2text dataset [36]. For all of these datasets, we use the standard evaluation protocol provided by the authors. Statistics of these datasets are given as follows.

Table 1: Comparison to the state-of-the-art on UCF101 and HMDB51 datasets (over all three splits), using just RGB modality.

| Method | Pre-training | Dataset | |
|---|---|---|---|
| | | UCF101 (%) | HMDB51 (%) |
| I3D [17] | ImageNet | 84.5 | 49.8 |
| TSN [33] | ImageNet | 86.4 | 53.7 |
| DTPP [37] | ImageNet | 89.7 | 61.1 |
| Res3D [9] | Sports-1M | 85.8 | 54.9 |
| TSN [33] | ImageNet + Kinetics | 91.1 | - |
| I3D [17] | ImageNet + Kinetics | **95.6** | **74.8** |
| ResNeXt-101 [38] | Kinetics | 94.5 | 70.2 |
| ARTNet [15] | Kinetics | 93.5 | 67.6 |
| T3D[14] | Kinetics | 91.7 | 61.1 |
| $ECO_{En}$ | Kinetics | 94.8 | 72.4 |

Table 2: Comparing performance of ECO with state-of-the-art methods on the Kinetics dataset.

| | Val (%) | | Test (%) |
|---|---|---|---|
| Methods | Top-1 | Avg | Avg |
| ResNeXt-101 [38] | 65.1 | 75.4 | **78.4** |
| Res3D [9] | 65.6 | 75.7 | 74.4 |
| I3D-RGB [17] | – | – | 78.2 |
| ARTNet [15] | 69.2 | 78.7 | 77.3 |
| T3D [14] | 62.2 | – | 71.5 |
| $ECO_{En}$ | **70.0** | **79.7** | 76.3 |

Table 3: Comparison with state-of-the-arts on Something-Something dataset. Last row shows the results using both Flow and RGB.

| Methods | Val (%) | Test (%) |
|---|---|---|
| I3D by [3] | - | 27.23 |
| M-TRN [39] | 34.44 | 33.60 |
| $ECO_{En}Lite$ | **46.4** | **42.3** |
| $ECO_{En}Lite\{^{RGB}_{Flow}$ | **49.5** | **43.9** |

The comparison is restricted to approaches that take the raw RGB videos as input without further pre-processing, for instance, by providing optical flow or human pose. The term $ECO_{NF}$ describes a network that gets $N$ sampled frames as input. The term $ECO_{En}$ refers to average scores obtained from an ensemble of networks with $\{16, 20, 24, 32\}$ number of frames.

### 5.1 Benchmark Comparison on Action Classification

The results obtained with ECO on the different datasets are shown in Tables 1, 2, and 3 and compares them to the state of the art. For UCF-101, HMDB-51, and Kinetics, ECO outperforms all existing methods except I3D, which uses a much heavier network. On Something-Something, it outperforms the other methods with a large margin. This shows the strong performance of the comparatively simple and small ECO architecture.

Table 4: Runtime comparison with state-of-the-art approaches using Tesla P100 GPU on UCF101 and HMDB51 datasets (over all splits). For other approaches, we just consider one crop per sample to calculate the runtime. We reported runtime without considering I/O.

| Method | Inference speed (VPS) | UCF101 (%) | HMDB51 (%) |
|---|---|---|---|
| Res3D [9] | <2 | 85.8 | 54.9 |
| TSN [33] | 21 | 87.7 | 51 |
| EMV [27] | 15.6 | 86.4 | - |
| I3D [17] | 0.9 | 95.6 | 74.8 |
| ARTNet [15] | 2.9 | 93.5 | 67.6 |
| $ECO_{Lite-4F}$ | **237.3** | 87.4 | 58.1 |
| $ECO_{4F}$ | **163.4** | 90.3 | 61.7 |
| $ECO_{12F}$ | **52.6** | 92.4 | 68.3 |
| $ECO_{20F}$ | **32.9** | 93.0 | 69.0 |
| $ECO_{24F}$ | **28.2** | 93.6 | 68.4 |

Table 5: Accuracy and runtime of ECO and ECO Lite for different numbers of sampled frames. The reported runtime is without considering I/O.

| Model | Sampled Frames | Speed (VPS) | | Accuracy (%) | | | |
|---|---|---|---|---|---|---|---|
| | | Titan X | Tesla P100 | UCF101 | HMDB51 | Kinetics | Someth. |
| ECO | 4 | 99.2 | 163.4 | 90.3 | 61.7 | 66.2 | – |
| | 8 | 49.5 | 81.5 | 91.7 | 65.6 | 67.8 | 39.6 |
| | 16 | 24.5 | 41.7 | 92.8 | 68.5 | 69.0 | 41.4 |
| | 32 | 12.3 | 20.8 | 93.3 | 68.7 | 67.8 | – |
| ECO Lite | 4 | 142.9 | 237.3 | 87.4 | 58.1 | 57.9 | – |
| | 8 | 71.1 | 115.9 | 90.2 | 63.3 | – | 38.7 |
| | 16 | 35.3 | 61.0 | 91.6 | 68.2 | 64.4 | 42.2 |
| | 32 | 18.2 | 30.2 | 93.1 | 68.3 | – | 41.3 |

## 5.2 Accuracy-Runtime Comparison

The advantages of the ECO architectures becomes even more prominent as we look at the accuracy-runtime trade-off shown in Table 4 and Fig. 4. The ECO architectures yield the same accuracy as other approaches at much faster rates.

Previous works typically measure the speed of an approach in frames per second (fps). Our model with ECO runs at 675 fps (and at 970 fps with ECO Lite) on a Tesla P100 GPU. However, this does not reflect the time needed to process a whole video. This becomes relevant for methods like TSN and ours, which do not look at every frame of the video, and motivates us to report *videos per second (vps)* to compare the speed of video understanding methods.

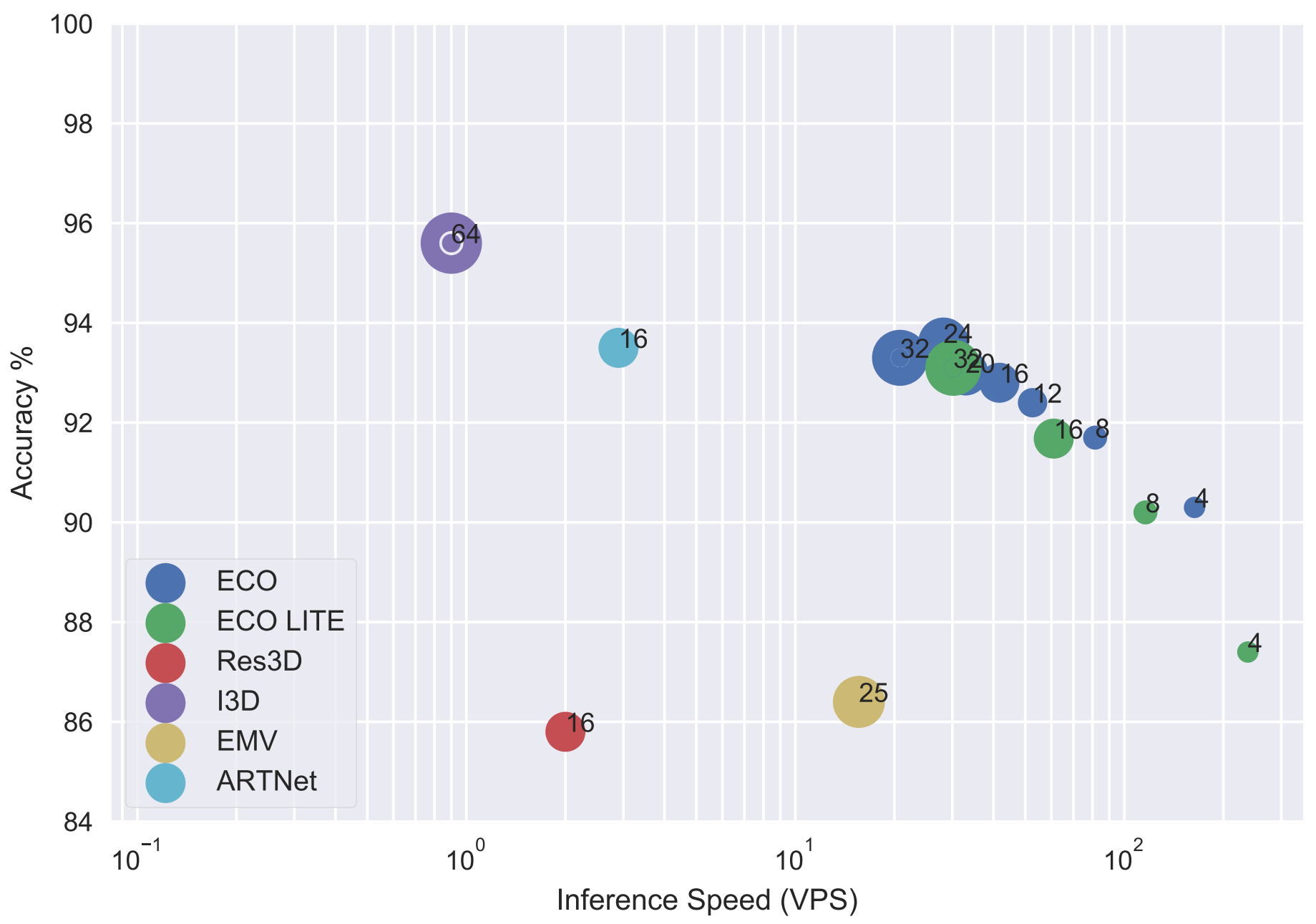

Fig. 4: Accuracy-runtime trade-off on UCF101 for various versions of ECO and other state-of-the-art approaches. ECO is much closer to the top right corner than any other approach.

ECO can process videos at least an order of magnitude faster than the other approaches, making it an excellent architecture for video retrieval applications.

**Number of sampled frames.** Table 5 compares the two architecture variants ECO and ECO Lite and evaluates the influence on the number of sampled frames $N$. As expected, the accuracy drops when sampling fewer frames, as the subsections get longer and important parts of the action can be missed. This is especially true for fast actions, such as "throw discus". However, even with just 4 samples the accuracy of ECO is still much better than most approaches in literature, since ECO takes into account the relationship between these 4 instants in the video, even if they are far apart. Fig. 6 even shows that for simple short-term actions, the performance decreases when using more samples. This is surprising on first glance, but could be explained by the better use of the network's capacity for simple actions when there are fewer channels being fed to the 3D network.

**ECO vs. ECO Lite.** The full ECO architecture yields slightly better results than the plain ECO Lite architecture, but is also a little slower. The differences in accuracy and runtime between the two architectures can usually be compensated by using more or fewer samples. On the Something-Something dataset, where the temporal context plays a much bigger role than on other datasets (see Figure 5), ECO Lite performs equally well as the full ECO architecture even with the same number of input samples, since the raw processing of single image cues has little relevance on this dataset.

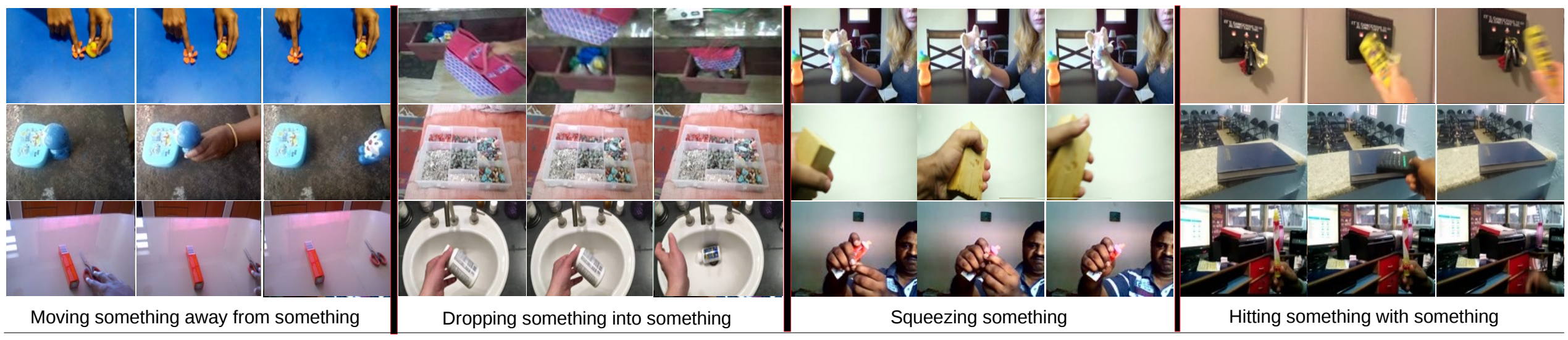

Fig. 5: Examples from the Something-Something dataset. In this dataset, the temporal context plays an even bigger role than on other datasets, since the same action is done with different objects, i.e., the appearance of the object or background gives almost no cues about the action.

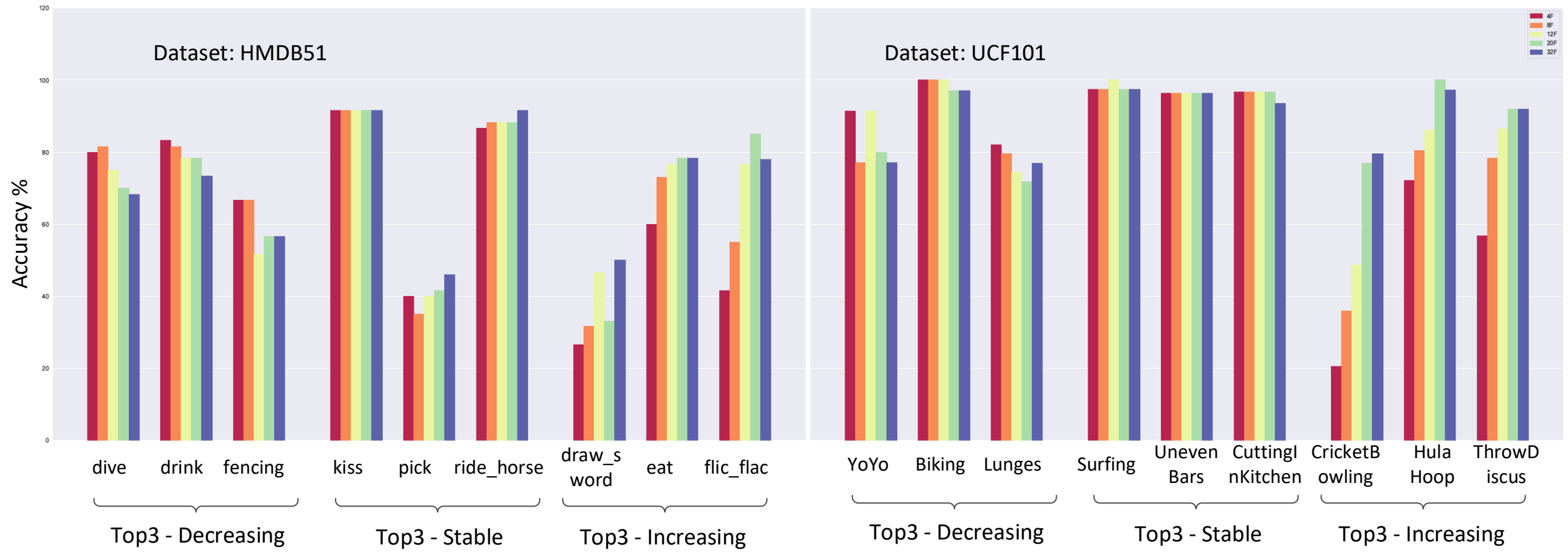

Fig. 6: Effect of the complexity of an action on the need for denser sampling. While simple short-term actions (leftmost group) even suffer from more samples, complex actions (rightmost group) clearly benefit from a denser sampling.

### 5.3 Early Action Recognition in Online Mode

Figure 7 evaluates our approach in online mode and shows how many frames the method needs to achieve its full accuracy. We ran this experiment on the J-HMDB dataset due to the availability of results from other online methods on this dataset. Compared to these existing methods, ECO reaches a good accuracy faster and also saturates at a higher absolute accuracy.

### 5.4 Video Captioning

To show the wide applicability of the ECO architecture, we also combine it with a video captioning network. To this end, we use ECO pre-trained on Kinetics to analyze the video content and train the state-of-the-art Semantic Compositional Network [25] for captioning. We evaluated on the the Youtube2Text (MSVD) dataset [36], which consists of 1,970 video clips with an average duration of 9 seconds and covers various types of videos, such as sports, landscapes, animals, cooking, and human activities. The dataset contains 80,839 sentences and each video is annotated with around 40 sentences.

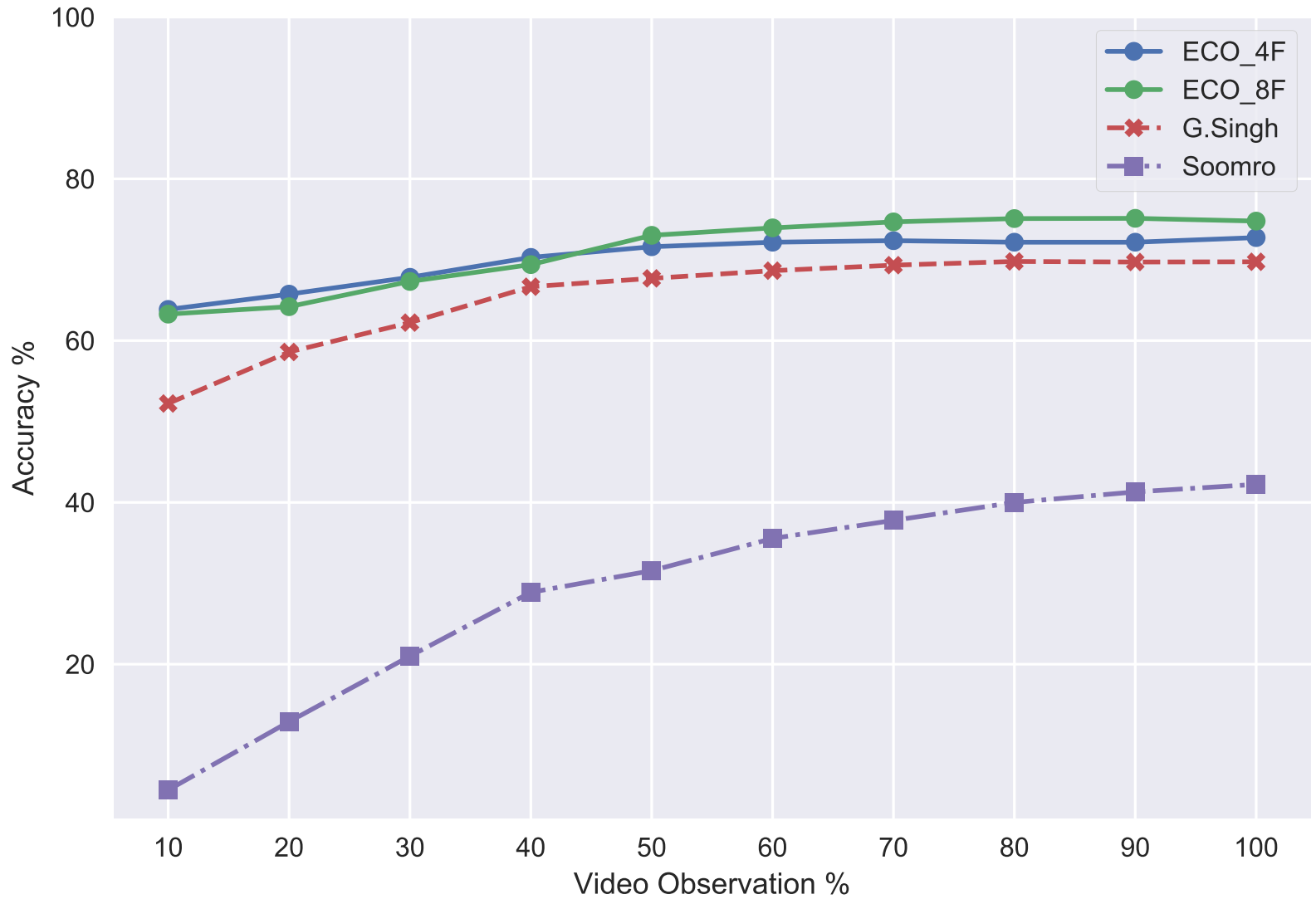

Fig. 7: Early action classification results of ECO in comparison to existing online methods [28,4] on the J-HMDB dataset. The online version of ECO yields a high accuracy already after seeing a short part of the video. Singh et al. [4] uses both RGB and optical flow.

Table 6 shows that ECO compares favorably to previous approaches across all popular evaluation metrics (BLEU[40], METEOR[41], CIDEr[42]). Even ECO Lite is already on-par with a ResNet architecture pre-trained on ImageNet. Concatenating the features from ECO with those of ResNet improves results further. Qualitative examples that correspond to the improved numbers are shown in Table 7.

## 6 Conclusions

In this paper, we have presented a simple and very efficient network architecture that looks only at a small subset of frames from a video and learns to exploit the temporal context between these frames. This principle can be used in various video understanding tasks. We demonstrate excellent results on action classification, online action classification, and video captioning. The computational load and the memory footprint makes an implementation on mobile devices a viable future option. The approaches runs 10x to 80x faster than state-of-the-art methods.

## Acknowledgements

We thank Facebook for providing us a GPU server with Tesla P100 processors for this research work.

Table 6: Captioning results on Youtube2Text (MSVD) dataset.

| | Metrics | | | |
|---|---|---|---|---|
| Methods | B-3 | B-4 | METEOR | CIDEr |
| S2VT [43] | - | - | 0.292 | - |
| GRU-RCN [44] | - | 0.479 | 0.311 | 0.678 |
| h-RNN [24] | - | 0.499 | 0.326 | 0.658 |
| TDDF [23] | - | 0.458 | 0.333 | 0.730 |
| AF [22] | - | 0.524 | 0.320 | 0.688 |
| SCN-c3d [25] | 0.587 | 0.482 | 0.330 | 0.692 |
| SCN-resnet [25] | 0.602 | 0.506 | 0.336 | 0.755 |
| SCN-Ensemble of 5 [25] | – | 0.511 | 0.335 | 0.777 |
| $\text{ECO}_{Lite-16F}$ | 0.601 | 0.504 | 0.339 | 0.833 |
| $\text{ECO}_{32F}$ | 0.616 | 0.521 | 0.345 | 0.857 |
| $\text{ECO}_{32F} + resnet$ | **0.626** | **0.535** | **0.350** | **0.858** |

Table 7: **Qualitative Results on MSVD.** First row corresponds to the examples where ECO improved over SCN and the second row shows the examples where ECO decreased the quality compared to SCN. $\text{ECO}_L$ refers to $\text{ECO}_{Lite-16F}$, ECO to $\text{ECO}_{32F}$, and $\text{ECO}_R$ to $\text{ECO}_{32F+resnet}$.

| | | |
|---|---|---|
| 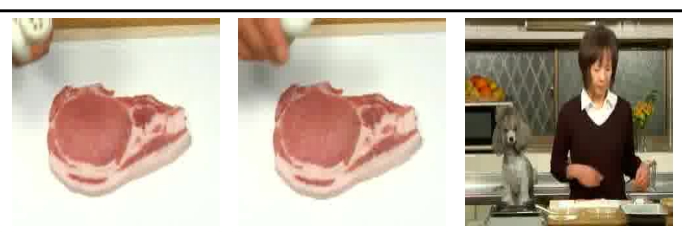 | 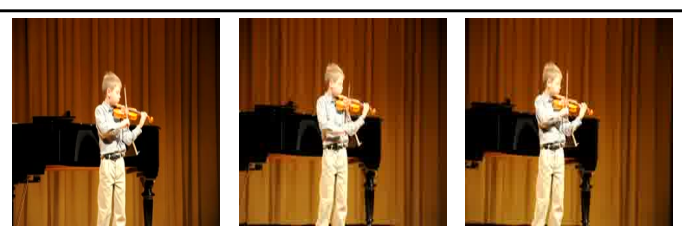 | 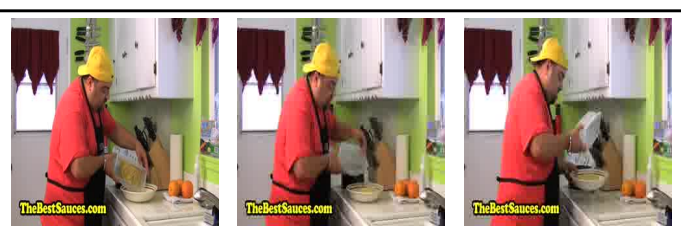 |
| **SCN**: a woman is cooking<br>**ECO$_L$**: the woman is seasoning the meat<br>**ECO**: a woman is seasoning some meat<br>**ECO$_R$**: a woman is seasoning some meat | **SCN**: a man is playing a flute<br>**ECO**: a man is playing a violin<br>**ECO$_R$**: a man is playing a violin<br>**ECO$_R$**: a man is playing a violin | **SCN**: a man is cooking<br>**ECO$_L$**: a man is pouring water into a container<br>**ECO**: a man is putting a lid on a plastic container<br>**ECO$_R$**: a man is draining pasta |
| 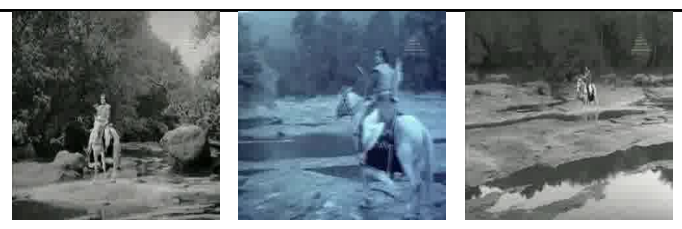 | 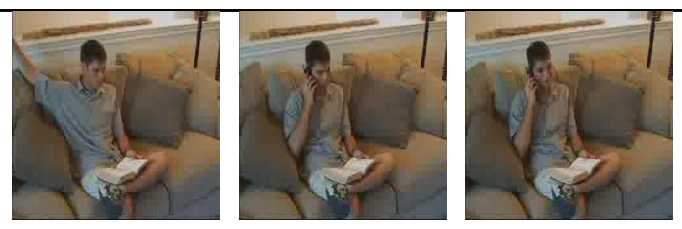 | 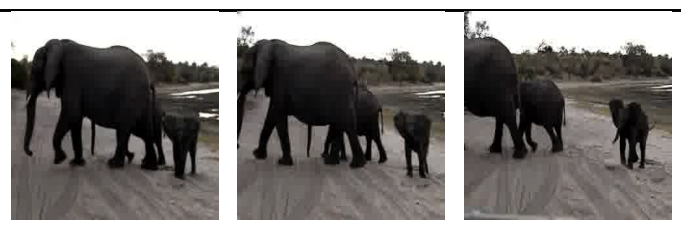 |
| **SCN**: a man is riding a horse<br>**ECO$_L$**: a woman is riding a motorcycle<br>**ECO**: a man is riding a horse<br>**ECO$_R$**: a man is riding a boat | **SCN**: a girl is sitting on a couch<br>**ECO$_L$**: a baby is sitting on the bed<br>**ECO**: a woman is playing with a toy<br>**ECO$_R$**: a woman is sleeping on a bed | **SCN**: two elephants are walking<br>**ECO$_L$**: a rhino is walking<br>**ECO**: a group of elephants are walking<br>**ECO$_R$**: a penguin is walking |

# –Supplementary Material–

Mohammadreza Zolfaghari, Kamaljeet Singh and Thomas Brox

University of Freiburg
{zolfagha,singhk,brox}@cs.uni-freiburg.de

**Abstract.** In this supplementary document we provide additional details and experimental results.

## 1 ECO

Figure 1 represents the architecture of our ECO model. In comparison to the ECO Lite model, ECO benefits from a 2D network in parallel to the 3D network that can directly provide visual semantics of individual frames. We apply average pooling for the 2D-Net$_s$ network to generate video-level features and then concatenate them with features obtained from 3D-net. The final output is a one-hot vector for the different class labels.

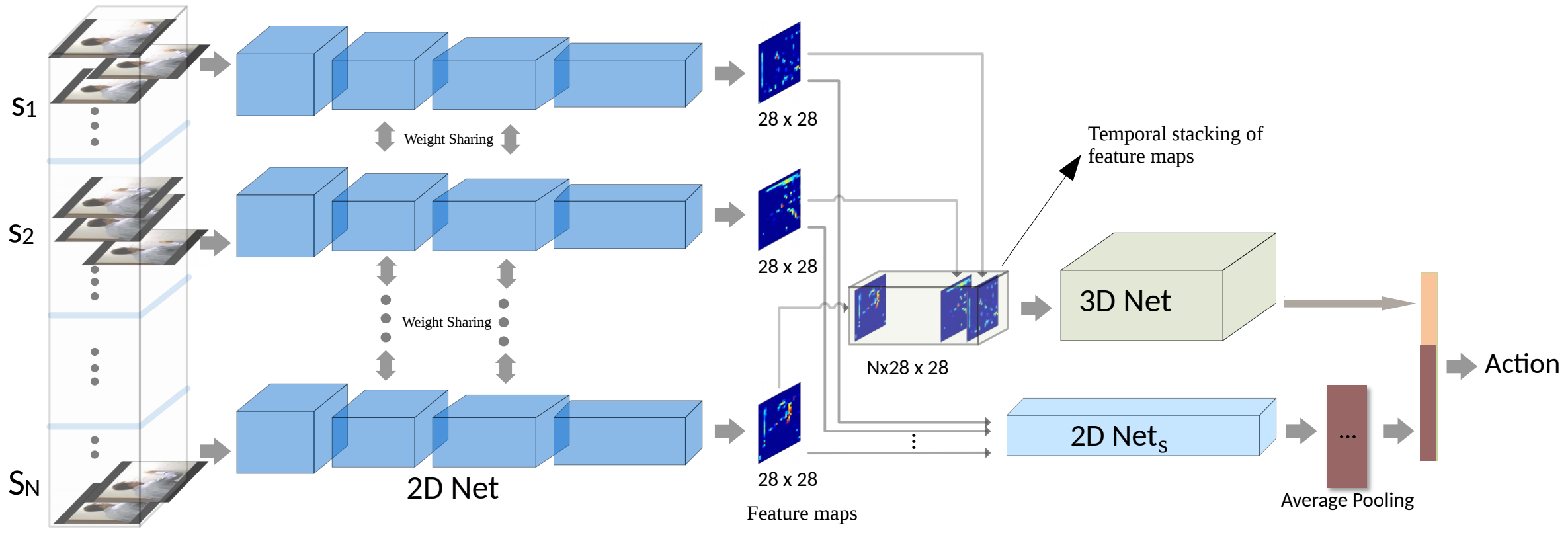

Fig. 1: **Architecture overview of ECO.** Each video is split into $N$ subsections of equal size. From each subsection a single frames is randomly sampled. The samples are processed by a regular 2D convolutional network to yield a representation for each sampled frame. In this design, we use a 2D network in parallel with the 3D network. 2D-$Net$ directly provides the visual semantics of single frames and 3D net processes the temporally stacked representations of frames using a 3D convolutional network. We apply average pooling for the 2D network to generate video-level features and concatenate them with the features from 3D-net. For simplicity, the figure just shows one channel of the 96 output channels of 2D-Net.

## 2 Network Architectures

Our ECO architecture consists of three submodules:

**2D-Net:** For the 2D network that exploits static semantics of individual frames, we use the first part of the BN-Inception architecture (until inception-3c layer) [1] as shown in Table 1. This network creates feature maps $M_i$ for $i^{th}$ input frame.

**3D-Net:** For the 3D network, we adopt several layers of 3D-Resnet18 [2], as show in Table 1. The concatenated output feature maps of 2D Net are fed as a single tensor $M_\varphi = [M_1, M_2, \cdots, M_N]$; $M_\varphi \in \mathbb{R}^{C \times N \times H \times W}$ to the 3D network, where $C$ is the number of channels at the last layer of 2D-Net, $N$ is the number of sampled frames, and $H = W = 28$ size of feature map (Fig. 1).

**2D-Net$_S$ :** In the ECO full design (Fig. 1), we use 2D-$Net_s$ in parallel with 3D-net. For this network, we use the BN-Inception architecture from the inception-4a layer before the last pooling layer [1].

Table 1: **Architecture details for 2D-Net and 3D-Net used in ECO**: The input to the network is $N$ frames of size $224 \times 224$.

| layer name | output size | 2D-Net ($H_{2D}$) | layer name | output size | 3D-Net ($H_{3D}$) |
|---|---|---|---|---|---|
| conv1_x | $112 \times 112$ | $\left[\text{2D conv } 7 \times 7 \text{ } 64\right]$ | conv3_x | $28 \times 28 \times N$ | $\begin{bmatrix}\text{3D conv } 3 \times 3 \times 3 \text{ } 128 \\ \text{3D conv } 3 \times 3 \times 3 \text{ } 128\end{bmatrix} \times 2$ |
| pool1 | $56 \times 56$ | $\left[\text{max pool } 3 \times 3\right]$ | conv4_x | $14 \times 14 \times \lfloor N/2 \rfloor$ | $\begin{bmatrix}\text{3D conv } 3 \times 3 \times 3 \text{ } 256 \\ \text{3D conv } 3 \times 3 \times 3 \text{ } 256\end{bmatrix} \times 2$ |
| conv2_x | $56 \times 56$ | $\left[\text{2D conv } 3 \times 3 \text{ } 192\right]$ | conv5_x | $7 \times 7 \times \lfloor N/4 \rfloor$ | $\begin{bmatrix}\text{3D conv } 3 \times 3 \times 3 \text{ } 512 \\ \text{3D conv } 3 \times 3 \times 3 \text{ } 512\end{bmatrix} \times 2$ |
| pool2 | $28 \times 28$ | $\left[\text{max pool } 3 \times 3\right]$ | | $1 \times 1 \times 1$ | pooling, "#c"-d fc, softmax |
| inception (3a) | $28 \times 28$ | $\left[- \text{ } 256\right]$ | – | – | – |
| inception (3b) | $28 \times 28$ | $\left[- \text{ } 320\right]$ | – | – | – |
| inception (3c) | $28 \times 28$ | $\left[- \text{ } 96\right]$ | – | – | – |

## 3 Sampling Function for Online Learning

For online video understanding, we use a strategy for sampling frames, which considers long-range information of the incoming stream while giving more importance to the more recent frames. We propose our sampling function as follows:

$$F_S^T = \{0.5^T Q_F^0\} \bigcup_{t=1}^{T} \{0.5^{(T-t+1)} Q_F^t\}, \tag{1}$$

where $Q_F$ is a queue of the last $N$ frames, $T = \lfloor i/N \rfloor - 1$ ; $i =$ frame number, and $N =$ number of samples. For instance, in the first time step, we will use all $N$ frames as input to the network:

$F_S^0 = \{0.5^0 Q_F^0\} = Q_F^0$

As an another example, at time step 2, we have:

$F_S^2 = \{0.5^2 Q_F^0\} \bigcup_{t=1}^{2} \{0.5^{(2-t+1)} Q_F^t\} = \{0.5^2 Q_F^0\} \bigcup \{0.5^2 Q_F^1\} \bigcup \{0.5^1 Q_F^2\}$,

i.e., current samples include 25% samples of $Q_F$ at time 0, 25% samples of $Q_F$ at time 1, and 50% of the last $N$ frames. As can be seen in Equation 1, recent frames contribute more than older frames. To avoid storing all incoming frames, we modify the strategy in a way that just keeps the sampled frames in memory:

$$F_S = \begin{cases} Q_F & \text{if } T = 0 \\ \{0.5\ Q_F\} \bigcup \{0.5\ S_F\} & \text{if } T > 0 \end{cases} \tag{2}$$

Where $S_F$ contains the sampled frames of the previous time step using $F_S$, and 0.5 means 50% of the samples. The function $F_S$ returns the sampled frames at each time $T$. The returned sampled frames are stored and updated incrementally in $S_F$ as explained in Algorithm 1, which allows us to keep only $S_F$ and $Q_F$ (queue of incoming N frames) in memory. As shown in Equation 2, at $T = 0$, $F_S$ just returns the first N frames, i.e., $Q_F$ but for $T > 0$ $F_S$ uniformly samples half of the frames from $Q_F$ and half from $S_F$.

The incremental updating of $S_F$ and sampling from the recent $Q_F$ frames ensures that the more recent frames are given more importance when fed into the proposed model, thereby making the model predictions more robust. Afterwards, the method feeds the sampled frames to ECO and updates the prediction by averaging the scores with the previous sampling and with the current sampling.

# 4 Video Length VS Number of Samples

In this experiment, we evaluate the effect of an increasing number of samples based on the video length. Therefore, we cluster videos by length into five categories [0-60], [60-120], [120-180], [180-240], and [240-320]. As shown in Fig. 2, action recognition on the short videos (length less than 60 frames) is harder task and sparse sampling limits the confusion. For longer videos, dense sampling helps up to some point.

# 5 Effect of Sampling Location

We evaluate the effect of the sampling location during test time. At inference time, we sample $N$ frames from the entire video with equal distances. In this experiment, we shift the location of samples temporally and present the results in terms of mean and standard deviation. Table 2 clearly shows that shifting the sampled location does not affect the performance excessively. In addition, an increasing number of samples decreases the standard deviation.

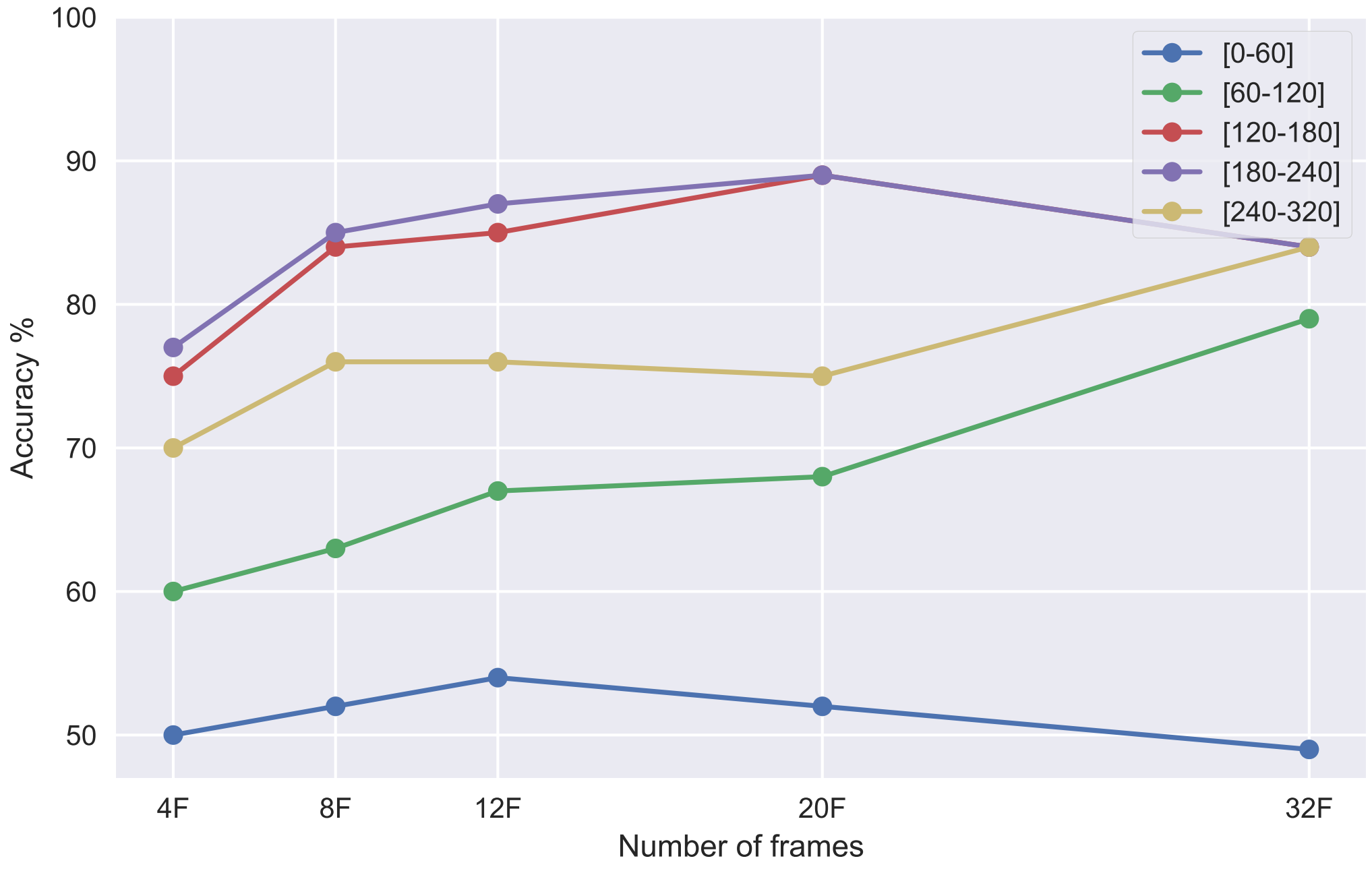

Fig. 2: Effect of increasing number of samples on accuracy for variable length of videos. For shorter videos, sparse sampling works better, while for longer videos dense sampling provides higher accuracy.

Table 2: Effect of the sampling location at inference time on the UCF101 and HMDB51 datasets (split1) using the ECO model.

| Datasets | Statistics | Number of Frames | | | | | |
|---|---|---|---|---|---|---|---|
| | | 4 | 8 | 12 | 16 | 24 | 32 |
| UCF101 | Mean | 89.83 | 91.81 | 92.42 | 92.73 | 93.30 | 92.09 |
| | Standard Deviation | 0.2329 | 0.0953 | 0.1525 | 0.1363 | 0.1189 | 0.1272 |
| HMDB51 | Mean | 62.62 | 65.88 | 69.67 | 68.91 | 69.42 | 69.48 |
| | Standard Deviation | 0.4460 | 0.6172 | 0.2106 | 0.5653 | 0.3339 | 0.3844 |

## 6 Early Action Recognition: UCF101

Fig. 3 evaluates the proposed method in the online learning mode. For this experiment we used split1 of the UCF101 dataset. As shown in Fig. 3, the approach performs already very well when just a few frames of the video have been observed.

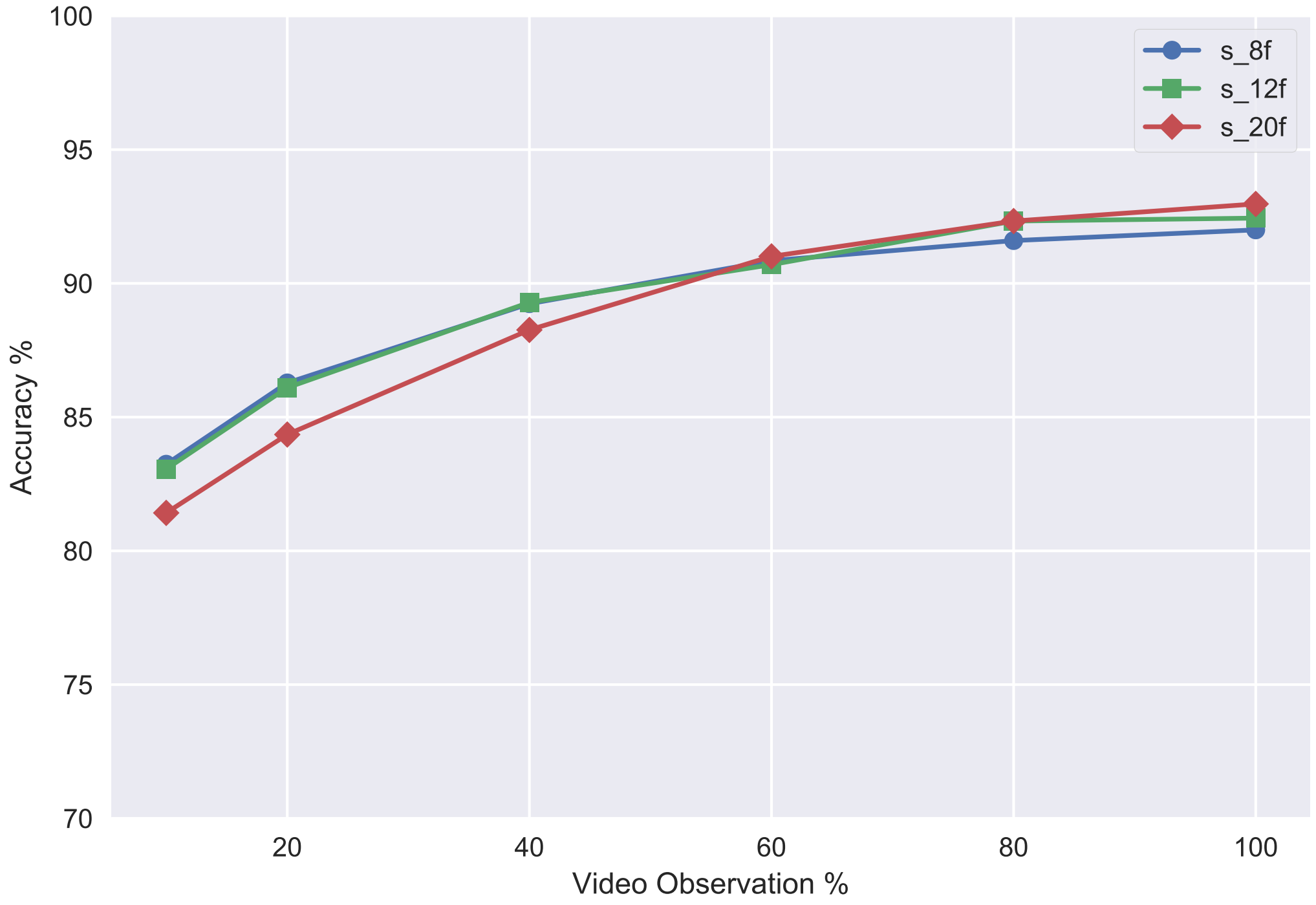

Fig. 3: Early action classification results of ECO on the UCF101 dataset (split1). ECO yields a high accuracy already after seeing a short part of the video.

# 7 More Qualitative Results on Video Captioning

Table 3 provides more qualitative results on the video captioning task. In this table, we compare the quality of captions produced by our approach to that of SCN [3]. ECO and SCN use the same language model for captioning, while the version using ECO benefits from the better feature representation of the video.

Table 3: Qualitative Results on MSVD, where ECO improved over SCN [3]. $\mathrm{ECO}_L$ refers to $\mathrm{ECO}_{Lite-16F}$, ECO to $\mathrm{ECO}_{32F}$, and $\mathrm{ECO}_R$ to $\mathrm{ECO}_{32F+resnet}$.

| | | |
|---|---|---|
| 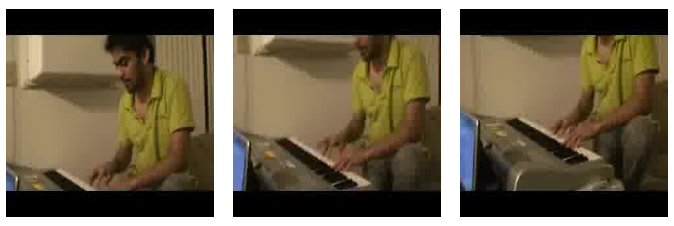 | 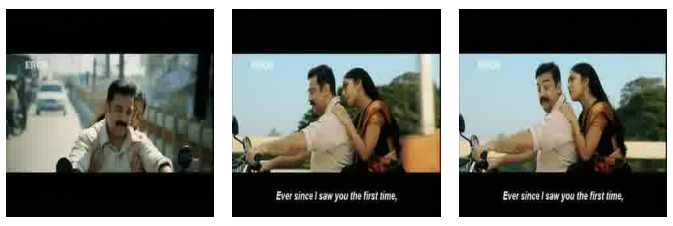 | 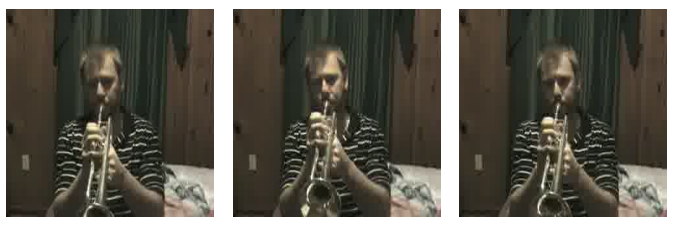 |
| **SCN**: a man is playing a guitar<br>**ECO**$_L$: a man is playing a keyboard<br>**ECO**: a man is playing a piano<br>**ECO**$_R$: a man is playing a piano | **SCN**: a man is singing<br>**ECO**$_L$: a man is riding a scooter<br>**ECO**: a man is riding a bike<br>**ECO**$_R$: a man is riding a bicycle | **SCN**: a boy is playing the music<br>**ECO**$_L$: a boy is playing a trumpet<br>**ECO**: a boy is playing a trumpet<br>**ECO**$_R$: a boy is playing a trumpet |
| 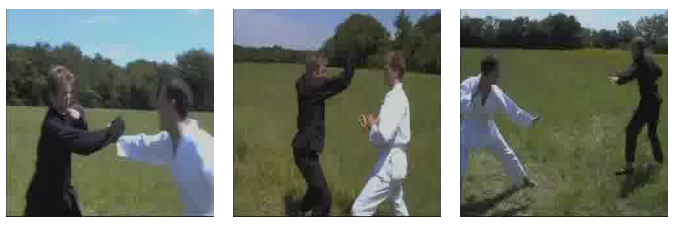 | 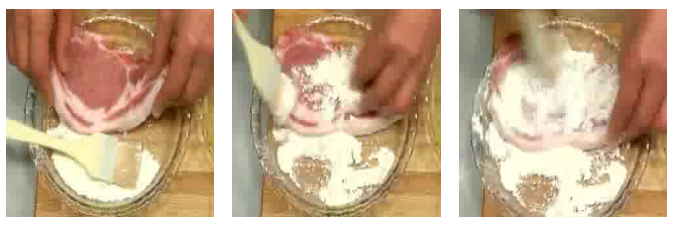 | 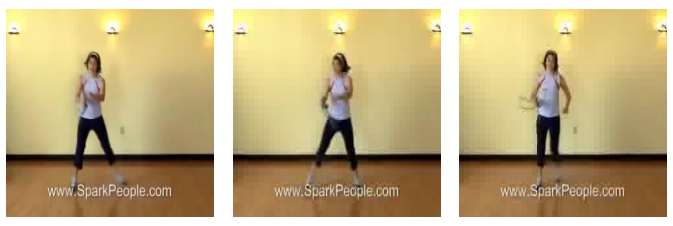  |
| **SCN**: a man is kicking a soccer ball<br>**ECO**$_L$: two men are fighting<br>**ECO**: a man is attacking a man<br>**ECO**$_R$: two men are fighting | **SCN**: a woman is mixing some meat<br>**ECO**$_L$: a woman is seasoning a piece of meat<br>**ECO**: a woman is mixing flour<br>**ECO**$_R$: a woman is coating flour | **SCN**: a boy is running<br>**ECO**$_L$: a boy is walking<br>**ECO**: a man is doing exercise<br>**ECO**$_R$: a man is exercising |